\documentclass{article} 
\usepackage{iclr2027_conference,times}

\usepackage{amsmath,amsfonts,bm}

\def\eqref#1{equation~\ref{#1}}

\def\1{\bm{1}}

\DeclareMathAlphabet{\mathsfit}{\encodingdefault}{\sfdefault}{m}{sl}
\SetMathAlphabet{\mathsfit}{bold}{\encodingdefault}{\sfdefault}{bx}{n}

\usepackage{hyperref}
\usepackage{url}
\usepackage{amsmath}
\usepackage{amssymb}
\usepackage{graphicx}
\usepackage{cleveref}
\usepackage{hyperref}
\usepackage{multirow}
\usepackage{booktabs}
\usepackage{makecell}
\usepackage{xspace} 
\usepackage{makecell}
\usepackage{tcolorbox}
\usepackage{svg}
\usepackage{caption}
\usepackage{pifont}
\usepackage[table]{xcolor}
\usepackage{colortbl}

\newcommand{\ourmodel}{LIFT\xspace}
\newcommand{\cmark}{\textcolor{green!60!black}{\ding{51}}}
\newcommand{\xmark}{\textcolor{red!70!black}{\ding{55}}}

\usepackage{enumitem}

\definecolor{firstcolor}{RGB}{245,190,190}   
\definecolor{secondcolor}{RGB}{243,211,170}  
\definecolor{thirdcolor}{RGB}{243,239,170}   
\definecolor{refgray}{RGB}{230,230,230}

\newcommand{\first}[1]{\cellcolor{firstcolor}{#1}}
\newcommand{\second}[1]{\cellcolor{secondcolor}{#1}}
\newcommand{\third}[1]{\cellcolor{thirdcolor}{#1}}

\title{LIFT: \textbf{L}ayout-\textbf{I}n-\textbf{F}u\textbf{t}ure Video Generation\\
under Large Viewpoint Change via On-Policy Self-Distillation}

\author{%
\textbf{Shengxiang Ji$^{1}$,
Boyang Wang$^{2}$,
Haiyang Xu$^{1}$,
Bingnan Li$^{1}$,
Yucheng Mao$^{1}$,
Zeyuan Chen$^{1}$,} \\
\textbf{Xiaojun Shan$^{1}$,
Xiang Zhang$^{3}$, 
Gang Hua$^{4}$,
Jianwen Xie$^{5}$,
Zezhou Cheng$^{2}$,
Zhuowen Tu$^{1}$} \\
$^{1}$UC San Diego
$^{2}$University of Virginia
$^{3}$Meta
$^{4}$Amazon
$^{5}$Lambda \\
\texttt{Project page: https://jsxzs.github.io/LIFT/} \\
}

\iclrfinalcopy 

\begin{document}
\maketitle
\fancyhead{}

\begin{center}
    \vspace{-2em}
    \centering
    \includegraphics[width=\linewidth]{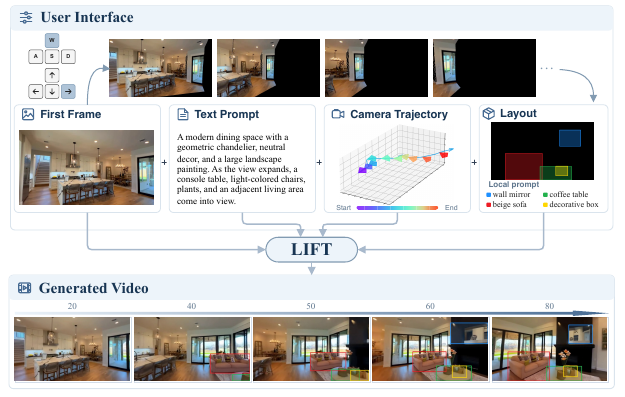}
    \vspace{-2em}
    \captionof{figure}{
        Given a first frame, users can navigate from the first-frame view along a desired camera path and specify layouts using bounding boxes with local text prompts in the final frame. Then, \textbf{\ourmodel} generates the intended shot that transitions from the input image to the user-defined last-frame layout following the prescribed camera trajectory.
    }
    \label{fig:teaser}
\end{center}

\begin{abstract}
\vspace{-0.5em}
We introduce \textbf{\ourmodel}, a unified image-to-video generation framework that complements camera control with \textbf{L}ayout-\textbf{I}n-\textbf{F}u\textbf{T}ure control, enabling users to specify what should appear in a future view and where it should appear.
This addresses a practical need in controllable video generation: given an initial image, users often care not only about how the camera moves, but also about what the scene should look like at key future moments, especially the final frame.
Existing camera controls specify viewpoint trajectories, while text prompts provide only coarse semantic guidance; neither precisely determines the content and spatial layout of future views.
This limitation becomes particularly pronounced under large viewpoint changes, where the camera reveals regions that are not visible in the first frame.
\ourmodel therefore uses the last-frame layout as an explicit control signal for the desired future scene.
Since learning from such sparse layout guidance is substantially more challenging than conditioning on dense per-frame layouts, we introduce on-policy self-distillation (OPSD) to transfer the control capability of a dense-layout teacher to a last-frame-layout student.
We further curate \textbf{LIFT-Vista}, a dataset featuring large viewpoint changes with camera and temporally consistent layout annotations.
Experiments show that \ourmodel improves video quality, future-layout controllability, and camera controllability over other methods.
\end{abstract}

\section{Introduction}
\label{sec:introduction}

Recent advances in video generation foundation models~\citep{wan2025wan,hacohen2026ltx,seedance2026seedance} have greatly improved the ability to synthesize high-fidelity, temporally coherent videos from text prompts or a single reference image.
Yet precise controllability remains a major barrier to using these models as practical creative tools, especially when the desired camera motion extends far beyond the initial view.
As illustrated in \cref{fig:teaser}, a creator may want the camera to move past the dining table and turn toward an unseen living room, while also specifying its composition—for example, a sofa facing the camera, a round coffee table in front of it, and a mirror above the fireplace.
Although these elements are not visible in the input image, their content and spatial layout determine what the newly revealed view should look like.

Existing controllable video generation methods address only part of this problem.
Camera-controlled video generation~\citep{he2024cameractrl,bai2025recammaster,robbyantteam2026lingbotworld} conditions on a prescribed camera trajectory to determine \textbf{how the viewpoint should move}.
However, when large camera motion reveals substantial regions outside the reference view, their content remains unspecified: the camera trajectory alone cannot determine \textbf{what should appear or where it should be placed}.
Layout guidance offers a natural complementary control by explicitly specifying the semantic content and spatial composition of such future views.

While layout-conditioned generation has been extensively studied for images~\citep{zhang2025creatilayout,huang2025laytrol}, it remains far less explored for video.
Existing video methods~\citep{li2025magicmotion,feng2025blobgenvid} typically rely on dense per-frame boxes, masks, or trajectories, and primarily focus on controlling the motion of objects already visible in the first frame.
Moreover, such dense frame-wise guidance places a substantial annotation burden on users.

To address these problems, we introduce \textbf{L}ayout-\textbf{I}n-\textbf{F}u\textbf{T}ure (LIFT), a unified video generation framework for large viewpoint changes that supports both camera control and last-frame layout conditioning.
\ourmodel operates in two inference modes: a \emph{single-condition mode}, conditioned only on the camera trajectory, and a \emph{dual-condition mode}, conditioned on both the camera trajectory and the last-frame layout.
As shown in \cref{fig:teaser}, \ourmodel enables users to control not only how the camera moves, but also what should appear in newly revealed regions and where it should appear.

Learning from only a last-frame layout is challenging: without dense per-frame layout guidance, the model must infer how specified objects evolve with camera motion and how the observed scene transitions toward the target composition.
We find that direct training with last-frame-only layouts under standard supervised flow matching struggles to exploit the sparse layout condition, leading to inferior future-layout control, while progressively reducing layout density through SFT incurs substantial training cost with limited gains.
We therefore first train a dense-layout model and use it as a teacher to supervise the last-frame-layout student on its own rollout states through on-policy self-distillation (OPSD)~\citep{zhao2026opsdllm,jiang2026dopsd,li2026rethinkingclassifierfreeguidanceonpolicy}.
Experiments demonstrate that OPSD achieves stronger layout control with fewer training sample updates than the SFT baselines.
Moreover, camera and layout conditioning are inherently coupled, as dense layouts also capture scene evolution induced by camera motion.
To exploit this coupling, we train a shared student across both the single-condition and dual-condition modes while distilling from the same dense-layout teacher.
This dual-mode training encourages the two forms of control to reinforce each other, improving both camera and future-layout controllability.

Our contributions are summarized as follows:
\begin{itemize}[leftmargin=1.5em,itemsep=0.5pt]
    \item We introduce \textbf{\ourmodel}, a unified video generation framework. It enables users to control both camera motion and the semantic-spatial composition of newly revealed regions using only a last-frame layout.

    \item We introduce dual-mode OPSD to this task, using dense spatiotemporal layouts as privileged information to train a shared student in both single-condition and dual-condition modes.

    \item We curate \textbf{LIFT-Vista}, a dataset tailored to large viewpoint changes.
    Our automatic pipeline identifies videos with substantial future-region revelation and produces temporally consistent camera and layout annotations.
\end{itemize}

\begin{figure*}[t]
    \centering
    \includegraphics[width=\linewidth]{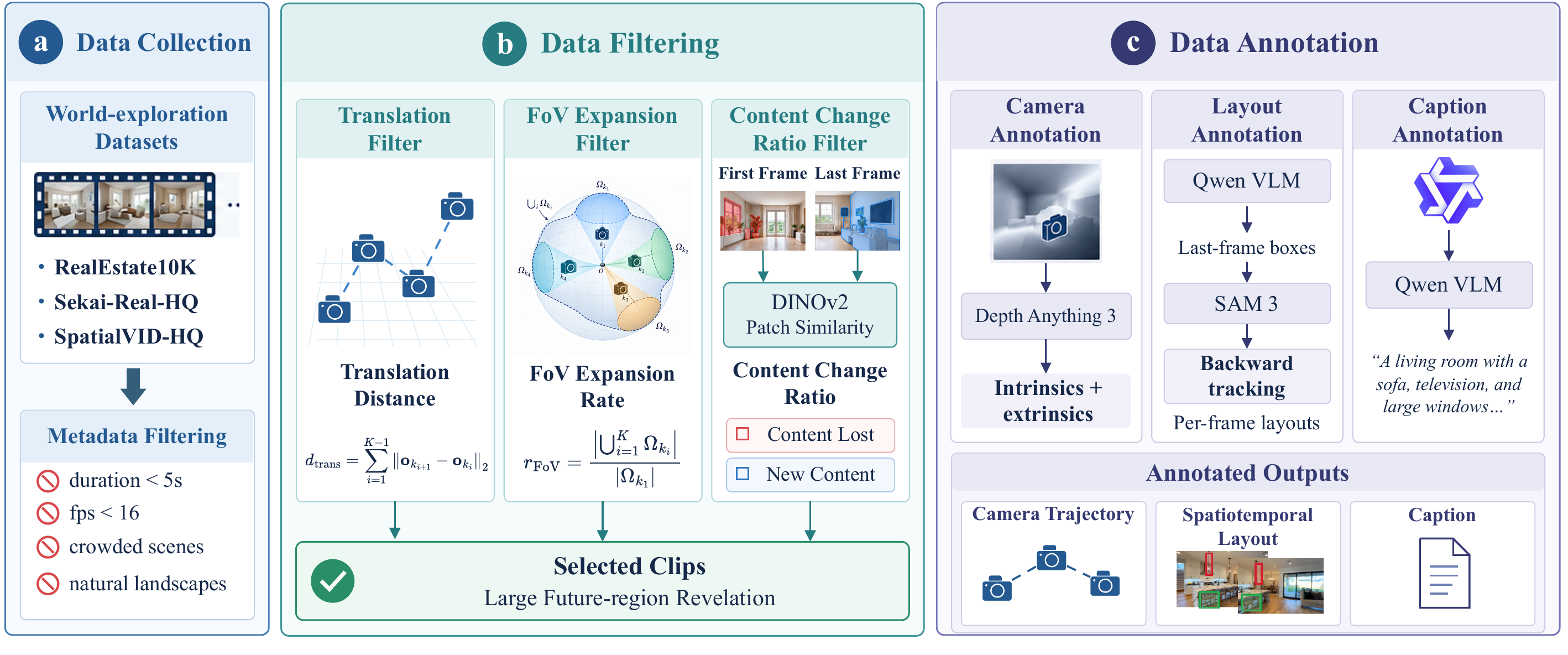}
    \caption{
    \textbf{Data Curation Pipeline.}
    (a) World-exploration video collection and metadata filtering.
    (b) Clip selection based on translation distance, FoV expansion, and content change.
    (c) Annotation of camera trajectories, spatiotemporal layouts, and captions.
    }
    \label{fig:data-curation}
\end{figure*}

\section{Data: LIFT-Vista}
\label{sec:data-curation}
Existing datasets don't directly support our target setting.
Camera-annotated video datasets~\citep{zhou2018realestate10k, li2026sekai, wang2025spatialvid} generally lack object-level layout labels, whereas datasets with bounding-box or layout~\citep{li2025magicmotion} typically lack camera trajectories and focus on the first-frame objects. We therefore curate LIFT-VISTA, a dataset specifically for future-view layout control under large viewpoint changes. Our data curation pipeline is illustrated in Fig.~\ref{fig:data-curation}.

\textbf{Data Collection.} We build LIFT-VISTA from RealEstate10K \citep{zhou2018realestate10k}, Sekai ~\citep{li2026sekai}, and SpatialVID ~\citep{wang2025spatialvid}.
The resulting data jointly provides camera trajectories and spatiotemporal object layouts, with an emphasis on scenes in which camera motion reveals regions outside the initial view. 

\textbf{Data Filtering.}
We first remove clips with undesirable scene properties, such as crowded scenes and natural landscapes.
We then retain clips with substantial future-region revelation using three complementary metrics: FoV expansion ratio, accumulated translation, and content change ratio.

We uniformly sample $K$ keyframes from each clip. 
For the $i$-th keyframe, let $\Omega_{k_i}\subseteq\mathbb{S}^2$ denote the set of visible viewing directions in a common world coordinate system. We estimate its spherical area using uniformly sampled directions on the unit sphere.
The FoV expansion ratio is defined as 
\begin{equation}
r_{\mathrm{FoV}} =
\frac{
\left|
\bigcup_{i=1}^{K}
\Omega_{k_i}
\right|
}{
|\Omega_{k_1}|
},
\end{equation}
which measures the total viewing region covered by the clip relative to the first frame,
and the accumulated camera translation as
\begin{equation}
d_{\mathrm{trans}} =
\sum_{i=1}^{K-1}
\left\|
\mathbf{o}_{k_{i+1}} - \mathbf{o}_{k_i}
\right\|_2,
\end{equation}
where $\mathbf{o}_{k_i}$ denotes the camera origin of the $i$-th keyframe.
Since camera motion alone does not directly measure changes in visible scene content, we additionally compute a patch-level CCR between the first and last frames using DINOv2~\citep{oquab2023dinov2}.
Let $\{\mathbf{p}_i\}_{i=1}^{N}$ and $\{\mathbf{q}_j\}_{j=1}^{N}$ denote their $\ell_2$-normalized patch embeddings.
The last-frame CCR is
\begin{equation}
r_{lf\text{-}CCR}
=
\frac{1}{N}\sum_{j=1}^{N}
\mathbb{I}\!\left[
\max_i \langle \mathbf{p}_i,\mathbf{q}_j\rangle < \tau
\right],
\label{eq:content-change-ratio}
\end{equation}
where $\mathbb{I}\!\left[\cdot\right]$ denotes the indicator function,
$\langle \cdot, \cdot \rangle$ denotes the cosine similarity, and $\tau$ is a similarity threshold. $r_{lf\text{-}CCR}$ measures the fraction of last-frame patches unmatched in the first frame.
We analogously compute $r_{ff\text{-}CCR}$ in the reverse direction to measure content leaving the initial view.

\textbf{Data Annotation.}
For camera trajectories, we apply Depth Anything 3~\citep{lin2025depth} to annotate camera intrinsics and extrinsics across all datasets, unifying coordinate systems. For spatiotemporal layout annotation, we first detect and annotate object-level bounding boxes in the last frame of each clip. Then, we use SAM3~\citep{carion2025sam} to track these objects throughout the entire clip, producing dense per-frame layouts.


\section{Method: \ourmodel}
\label{sec:method}

\subsection{Preliminary}


\textbf{Diffusion On-Policy Self-Distillation.}
OPSD uses the same model to act as both student and teacher. The student is conditioned only on the inference-time context $c$, whereas the teacher additionally observes privileged information $r$. In the LLM domain, the student is trained to match the teacher distribution using reverse KL. Recent works~\citep{fang2026flowopd, li2026diffusionopd,zhou2026danceopd} study on-policy distillation for diffusion models. In our ODE-based rollout setting, we use the following velocity-matching surrogate objective:
\begin{equation}
\mathcal{L}_{\mathrm{OPSD}}^{\mathrm{FM}}(\theta)
=
\mathbb{E}_{x_{t_0:t_N}\sim p_\theta(\cdot\mid c)}
\left[
\sum_{j=0}^{N-1}
w(t_j)
\left\|
v_{\theta}(x_{t_j}, t_j, c)
-
\operatorname{sg}\!\left[v_{\theta_{old}}(x_{t_j}, t_j, c, r)\right]
\right\|_2^2
\right].
\end{equation}
where $\theta_{\mathrm{old}}$ denotes the frozen teacher parameters, $w(t_j)$ is an optional timestep-dependent weighting function, and $\operatorname{sg}[\cdot]$ denotes the stop-gradient operation.

\subsection{Joint Camera and Layout Conditioned DiT}

\begin{figure*}[t]
    \centering
    \vspace{-1em}
    \includegraphics[width=0.9\linewidth]{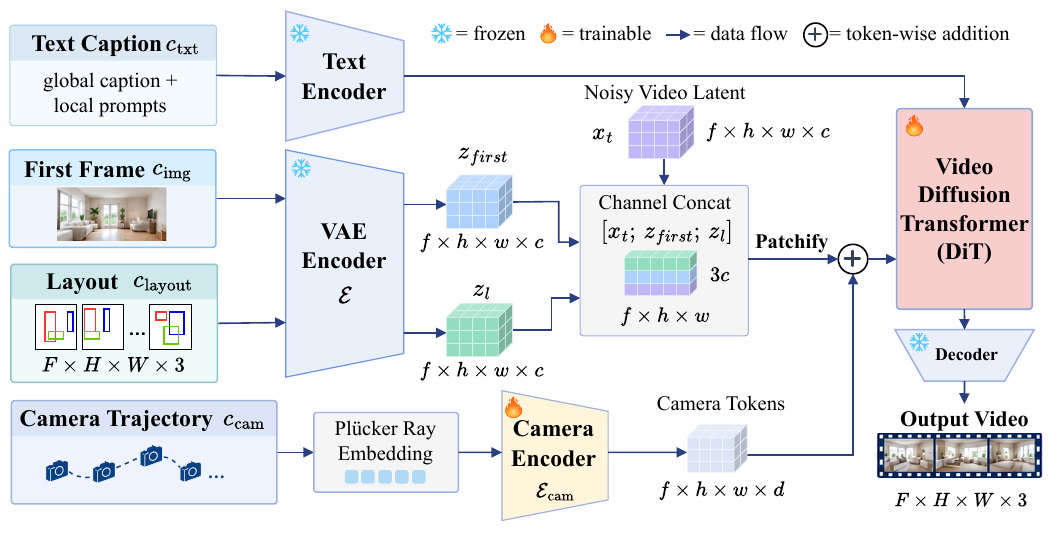}
    \vspace{-1em}
    \caption{
        \textbf{Model Architecture.} Layout latent is channel-concatenated with the noisy video latent and first-frame latent.
        Camera tokens are injected into the DiT stream through token-wise addition.
        Color-referenced layout local prompts are appended to the global caption.
    }
    \label{fig:arch}
    \vspace{-1em}
\end{figure*}

We build an image-to-video diffusion model jointly conditioned on four signals: a reference first frame $c_{\mathrm{img}}$ that specifies the initial scene appearance, a text caption $c_{\mathrm{txt}}$ describing the video content, a target camera trajectory $c_{\mathrm{cam}}$ specifying the viewpoint change, and a spatiotemporal layout $c_{\mathrm{layout}}$ specifying the locations and semantics of objects at the conditioned frames.
An overview of the architecture is shown in \cref{fig:arch}.

\textbf{Layout Control.}
We introduce layout maps to explicitly control layouts throughout the generated video.
Using the layout annotations described in Sec.~\ref{sec:data-curation}, we render the bounding boxes into a pixel-aligned layout map video \(m_{l}\), where each object instance is assigned a unique color that remains consistent across frames to preserve its identity.
The layout map is encoded by the shared VAE encoder \(\mathcal{E}\):
$z_{l} = \mathcal{E}(m_{l})$.
We then concatenate the noisy video latent \(x_t\), the first-frame latent \(z_{\mathrm{first}}\), and the layout latent $z_l$ along the channel dimension:
\begin{equation}
\tilde{x}_t =
\operatorname{Concat}_{\mathrm{ch}}
\left(x_t, z_{\mathrm{first}}, z_l\right),
\label{eq:latent_concat}
\end{equation}
where $\tilde{x}_t$ is subsequently projected into visual tokens by the
patchification layer.
The layout map specifies \emph{where} objects should appear, while their
semantic information is provided through text.
Specifically, we associate each object description with its corresponding
bbox color and append these local object prompts to the global video caption.
Together, the layout map and color-referenced local prompts provide geometric and semantic control.

\begin{figure*}[t]
    \centering
    \vspace{-1em}
    \includegraphics[width=\linewidth]{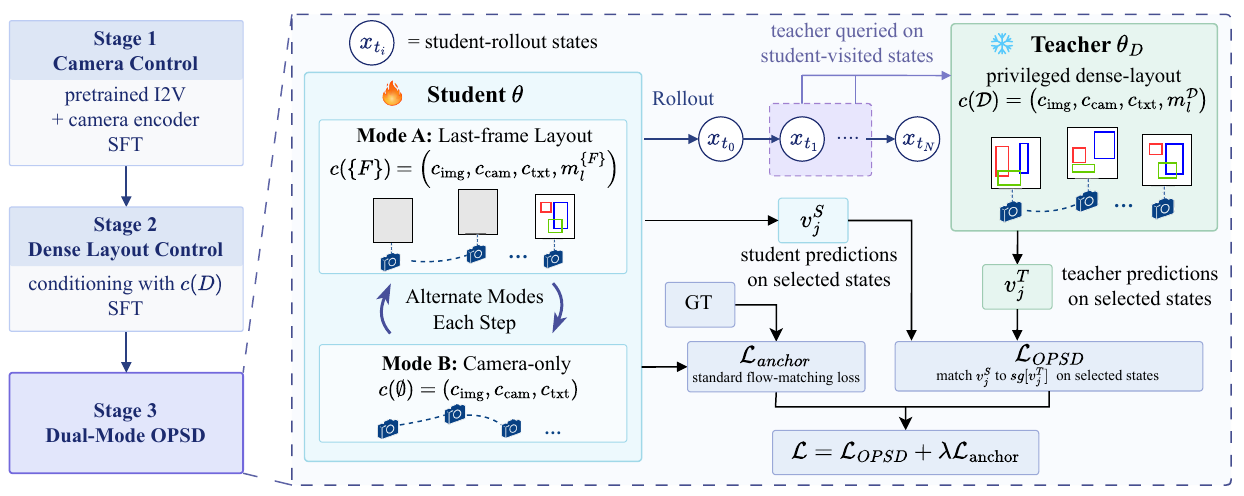}
    \vspace{-1.5em}
    \caption{
        \textbf{Dual-mode OPSD.} The student alternates between last-frame-layout and camera-only modes and distills selected rollout states from a frozen dense-layout teacher with a flow-matching anchor loss. The student and teacher are both initialized from $\theta_{\mathcal{D}}$.
    }
    \label{fig:opsd}
    \vspace{-1em}
\end{figure*}

\textbf{Camera Control.} 
We adopt Pl\"ucker ray embeddings $\mathcal{P} \in \mathbb{R}^{F \times H \times W \times 6}$ ~\citep{he2024cameractrl, bahmani2025ac3d} as the camera representation, which provide strong per-pixel geometric information. A lightweight camera encoder $\mathcal{E}_{\mathrm{cam}}$ transforms the Pl\"ucker representation into camera latent tokens that are spatiotemporally aligned with the patchified video tokens~\citep{he2025cameractrl2,wan2025wan}.
These camera tokens are then injected into the DiT stream through token-wise addition:
\begin{equation}
    \mathcal{H}_{in} 
    =
    \operatorname{patchify}(\tilde{x}_t)
    +
    \mathcal{E}_{\mathrm{cam}}(\mathcal{P}),\end{equation}
where $\mathcal{H}_{\mathrm{in}}$ is fed into the diffusion Transformer.
This spatiotemporally aligned camera conditioning allows the denoising network to directly associate video contents with the prescribed camera motion.

\subsection{Dual-Mode OPSD Training}
\label{sec:training}
Our model needs to integrate two controls ---camera trajectory and future-view layout.
In particular, we find that directly learning last-frame-only layout
conditioning with standard SFT is highly challenging.
We therefore adopt OPSD to reach the final last-frame-layout regime, which is much more data-efficient and effective.

\textbf{Conditioning Modes.}
Let $\mathcal{S}\subseteq\{1,\dots,F\}$ denote the set of frames at which the layout is exposed to the model.
The layout map $m_{l}^{\mathcal{S}}$ renders object boxes only at frames in $\mathcal{S}$ and leaves other frames empty. 
The corresponding conditioning context is
\begin{equation}
c(\mathcal{S})
=
\left(
c_{\mathrm{img}},
c_{\mathrm{cam}},
c_{\mathrm{txt}},
m_l^{\mathcal{S}}
\right)
\label{eq:cond-mode}
\end{equation}
We define
$\mathcal{S}=\mathcal{D}\triangleq\{1,\ldots,F\}$ as the \emph{dense-layout} mode,
$\mathcal{S}=\{F\}$ as the \emph{lastframe-layout} mode (i.e. dual-condition mode), and $\mathcal{S}=\emptyset$ as the
\emph{camera-only} mode (i.e. single-condition mode).

Our training has 3 stages: camera control, dense layout control, and dual-mode OPSD. For stage~1, we train the camera controllability, adapting the model to our task setting (i.e. large viewpoint changes and future-region revelation) \cref{sec:data-curation}. For stage~2, we introduce the layout conditioning and continue SFT under the dense layout context $c(\mathcal{D})$. The resulting weights, denoted as $\theta_{\mathcal{D}}$, serve both as the teacher and as the student initialization for stage~3.

\textbf{Dual-Mode OPSD.}
Camera and layout control are not fully independent.
A dense spatiotemporal layout implicitly describes how the scene evolves under viewpoint changes and can therefore convey part of the camera-induced motion~\citep{li2025magicmotion,wang2025cinemaster}.
As illsustrated in \cref{fig:opsd}, we perform OPSD in two student conditioning modes: $\mathcal{S}\in \left\{ \{F\},\emptyset \right\}$, to jointly improve lastframe-layout control and camera-only control.
Both student modes share the same parameters and are distilled from the same dense-layout teacher. 

\textbf{OPSD Objective.}
We freeze the Stage~2 model $\theta_{\mathcal{D}}$ as the teacher and
initialize the student $\theta$ from the same parameters.
For each training sample, we first perform on-policy rollout
with the student under $c(\mathcal{S})$, without gradient tracking.
The frozen dense-layout teacher is then queried at selected states along the
student trajectory:
\begin{equation}
\begin{aligned}
\mathcal{L}_{\mathrm{OPSD}}(\theta;\mathcal{S})
&=
\mathbb{E}_{
x_{t_0:t_N}
\sim
p_{\theta}(\cdot\mid c(\mathcal{S}))
}
\left[
\frac{1}{|\mathcal{K}_{\mathcal{S}}|}
\sum_{j\in\mathcal{K}_{\mathcal{S}}}
w(t_j)
\left\|
v_j^{S}
-
\operatorname{sg}[v_j^{T}]
\right\|_2^2
\right],
\end{aligned}
\label{eq:opsd-loss}
\end{equation}
where $x_{t_0:t_N}$ denotes the student rollout trajectory, 
$v_j^{S}
=
v_{\theta}(x_{t_j},t_j,c(\mathcal{S}))$
and
$v_j^{T}
=
v_{\theta_{\mathcal{D}}}(x_{t_j},t_j,c(\mathcal{D}))$ denote the student and teacher velocity predictions, respectively, $\mathcal{K}_{\mathcal{S}}\subseteq\{0,\ldots,N-1\}$ denotes the
subset of student-visited states queried for distillation, and $\operatorname{sg}$ denotes stop-gradient.

\textbf{Anchoring Loss.}
Although dense layout provides the teacher with better spatiotemporal control, the teacher itself is imperfect. Optimizing the OPSD objective \cref{eq:opsd-loss} alone can degrade generation quality. Therefore, we maintain the standard flow-matching objective as an anchoring loss:
\begin{equation}
\mathcal{L}_{\mathrm{anchor}}(\theta;\mathcal{S})
=
\mathbb{E}_{x_0,\epsilon,t}
\left[
\left\|
v_\theta\bigl(x_t^{\mathrm{FM}},t,c(\mathcal{S})\bigr)
-
v_t^\star(x_0,\epsilon)
\right\|_2^2
\right].
\label{eq:anchor}
\end{equation}
where $v_t^\star(x_0,\epsilon)$ denotes the flow matching velocity target. 
This term helps maintain generation fidelity while OPSD transfers dense-layout knowledge to sparse conditioning modes.

\textbf{Full Objective.} 
The overall Stage~3 objective is
\begin{equation}
\mathcal{L}(\theta)
=
\mathbb{E}_{\mathcal{S}\sim\pi}
\Bigl[
\mathcal{L}_{\mathrm{OPSD}}(\theta;\mathcal{S})
+
\lambda\,\mathcal{L}_{\mathrm{anchor}}(\theta;\mathcal{S})
\Bigr],
\label{eq:total}
\end{equation}
where $\pi$ is the sampling distribution over the two target conditioning modes $\mathcal{S}\in\{\{F\},\emptyset\}$, and $\lambda$ is the anchor weight, where we set the anchor weight to $\lambda = 0.1$ for Stage 3 training.

\textbf{Selective State Distillation.}
Not all states along the student rollout provide equally useful distillation signals.
The global spatial layout configuration is largely determined during the early, high-noise stage of the denoising trajectory~\citep{hertz2022prompt}.
At these states, we observe that the teacher with privileged dense-layout conditioning can correct the student to the desired layout.
In contrast, at later low-noise states, the teacher produces nearly no corrections, making the corresponding distillation signal less informative.
We therefore concentrate OPSD supervision on the first 10 high-noise states of each student rollout.

\vspace{-0.5em}
\section{Experiment}

\subsection{Implementation Details}
\vspace{-0.25em}
We build our model on top of Wan2.1-Fun-V1.1-1.3B-Control-Camera~\citep{wan2025wan}.
All experiments are conducted at a resolution of $352\times640$, using 81-frame clips at 16 FPS.
Training is performed on 4 NVIDIA H100 GPUs.
For the three training stages, we optimize the model for 8,000, 4,000, and 500 steps, respectively.
The corresponding global batch sizes are 32, 32, and 16, with learning rates of $1\times10^{-5}$, $1\times10^{-4}$, and $5\times10^{-5}$.
We use AdamW as the optimizer.
For Stage~3, the sampling probabilities for the two OPSD modes are $P(\mathcal{S}=\{F\})=0.7$ and $P(\mathcal{S}=\emptyset)=0.3$ for the lastframe-layout and camera-only modes, respectively.
For inference, we use 50 denoising steps and a cfg scale of 6.0. 
More implementation details are included in \cref{supp:implem} and \cref{supp-sec:data-curation}.

\subsection{Quantitative and Qualitative Comparisons}
\vspace{-0.25em}
\textbf{Baselines.}
We compare against three categories of controllable video generation methods: camera control, object motion control, and joint camera-and-object motion control.
For camera control, we evaluate against two recent state-of-the-art methods, Uni3C~\citep{cao2025uni3c} and GEN3C~\citep{ren2025gen3c}.
For object motion control, we compare with MagicMotion~\citep{li2025magicmotion}, which uses bounding-box trajectories to specify object motion.
We further include Direct-a-Video~\citep{yang2024directavideo} as a joint-control baseline.
Direct-a-Video supports training-free control of object motion using bounding-box trajectories, whereas its camera control is restricted to horizontal/vertical panning and zooming.
For a fair comparison to baselines, we provide MagicMotion and Direct-a-Video with dense per-frame layout trajectories, whereas our model uses only a last-frame layout.

\begin{figure*}[!th]
    \centering
    \includegraphics[width=\linewidth]{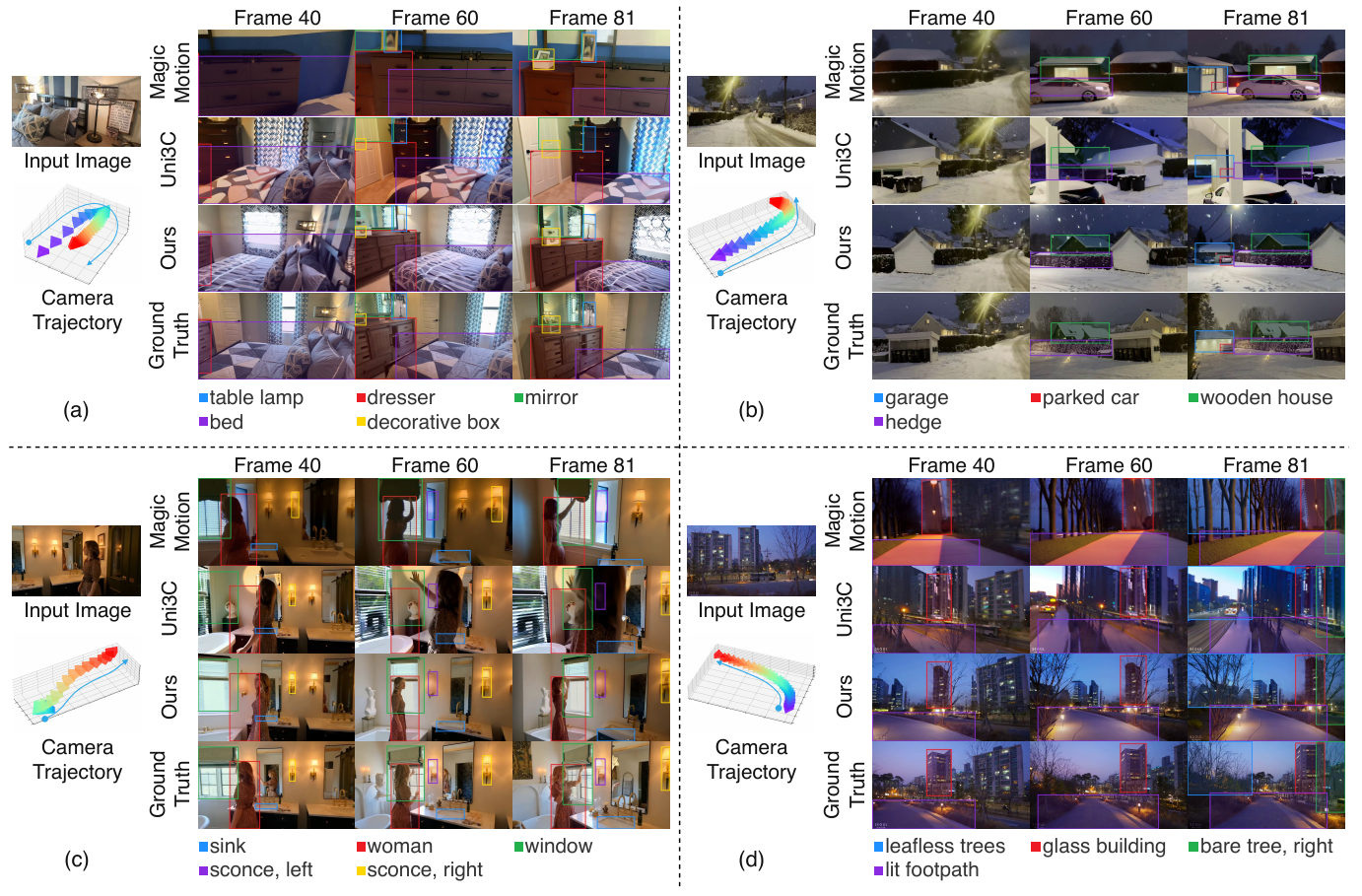}
    \vspace{-2em}
    \caption{
    \textbf{Qualitative comparison.}
    Bounding boxes indicate the target locations of conditioned objects.
    MagicMotion lacks explicit camera control and often produces inconsistent scene evolution or incorrect objects, while Uni3C follows the prescribed camera trajectory but leaves newly revealed regions uncontrolled, resulting in unspecified contents, e.g., a pillar in (b) and a car street in (d).
    In contrast, ours follows the prescribed camera motion and realizes the specified future-view layout.    
    }  
    \label{fig:qualitative}
\end{figure*}

\begin{table*}[!th]
\centering
\small
\setlength{\tabcolsep}{4pt}
\caption{
\textbf{Quantitative comparison} with state-of-the-art controllable video generation methods.
We report video quality, camera trajectory accuracy, and semantic consistency metrics.
The \protect\colorbox{firstcolor}{best}, 
\protect\colorbox{secondcolor}{second-best}, and 
\protect\colorbox{thirdcolor}{third-best} results are highlighted accordingly.
}
\vspace{-0.5em}

\resizebox{\textwidth}{!}{
\begin{tabular}{l|ccc|cccccccc}
\toprule
\multirow{2}{*}{\textbf{Method}}
& \multicolumn{3}{c|}{\textbf{Control}}
& \multicolumn{3}{c}{\textbf{Video Quality}}
& \multicolumn{2}{c}{\textbf{Camera Error}}
& \multicolumn{3}{c}{\textbf{Semantic Consistency}} \\
\cmidrule(lr){2-4}
\cmidrule(lr){5-7}
\cmidrule(lr){8-9}
\cmidrule(lr){10-12}
&
Camera
& \makecell{Object\\Motion}
& \makecell{Future\\Layout}
& FVD $\downarrow$
& FID $\downarrow$
& LPIPS $\downarrow$
& RotErr $\downarrow$
& TransErr $\downarrow$
& mIoU $\uparrow$
& $SR_e$ $\uparrow$
& $\mathrm{CLIP}_{\mathrm{local}}$ $\uparrow$ \\
\midrule

Direct-a-Video
& \cmark & \cmark & \xmark
& 539.28
& 71.56
& 0.80
& 28.48
& 2.28
& 0.10
& 0.11
& 0.14 \\

MagicMotion
& \xmark & \cmark & \xmark
& 277.94
& 18.32
& \third{0.57}
& 15.74
& \third{1.59}
& \second{0.41}
& \second{0.53}
& \second{0.21} \\

GEN3C
& \cmark & \xmark & \xmark
& \first{89.59}
& \third{15.60}
& \second{0.51}
& \third{3.91}
& 2.62
& 0.18
& 0.38
& 0.19 \\

Uni3C
& \cmark & \xmark & \xmark
& \third{111.60}
& \first{12.75}
& \first{0.42}
& \second{3.23}
& \second{0.70}
& \third{0.31}
& \third{0.47}
& \second{0.21} \\

\textbf{\ourmodel}
& \cmark & \cmark & \cmark
& \second{99.35}
& \second{12.84}
& \first{0.42}
& \first{2.97}
& \first{0.59}
& \first{0.51}
& \first{0.59}
& \first{0.24} \\

\bottomrule
\end{tabular}
}
\vspace{-1em}
\label{tab:quantitative_comparison}
\end{table*}

\textbf{Metrics.}
We evaluate generated videos in terms of visual quality, camera controllability,
and layout controllability.
For visual quality, we report FVD~\citep{unterthiner2018towards},
FID~\citep{heusel2017gans}, and LPIPS~\citep{zhang2018unreasonable}.
For camera control, we measure rotation error (RotErr) and translation error
(TransErr) between the generated and target camera trajectories (~\citep{zhang2025dualcamctrl}).
For layout control, following OverLayBench~\citep{li2026overlaybench}, we report
mIoU, entity success rate (SR$_e$), and
CLIP$_{\mathrm{local}}$~\citep{radford2021learning} to evaluate spatial
alignment, entity-level success, and local semantic consistency, respectively.

\textbf{Quantitative and Qualitative Results.}
As shown in Tab.~\ref{tab:quantitative_comparison}, our method achieves strong performance across all three evaluation dimensions.
\ourmodel (1.3B) achieves competitive video quality against 14B Uni3C and 7B GEN3C.
\ourmodel also achieves the best camera-control accuracy among the compared methods while supporting future-layout control.
More importantly, \ourmodel consistently achieves the best layout-control performance.
Compared with MagicMotion, which is additionally provided with dense per-frame bounding-box trajectories, \ourmodel improves mIoU from 0.41 to 0.51, despite requiring only a last-frame layout.
These results demonstrate that \ourmodel effectively combines camera control with future-view spatial control while maintaining competitive generation quality. We show more visualization qualitative results in \cref{fig:qualitative} and supplementary materials.

\subsection{Ablation Study}

\begin{table*}[!t]
\centering
\small
\setlength{\tabcolsep}{4.5pt}
\caption{
\textbf{Ablation study of SFT and OPSD.}
D2S-SFT denotes dense-to-sparse curriculum SFT.
Training cost is measured by the number of training sample updates (steps $\times$ global batch size).
OPSD achieves the best results with much fewer training samples of the SFT-based alternatives.
}
\vspace{-0.5em}

\resizebox{\textwidth}{!}{
\begin{tabular}{l|c|cccccccc}
\toprule
\multirow{2}{*}{\textbf{Method}}
& \multirow{2}{*}{\makecell[c]{\textbf{Training Sample}\\\textbf{(Steps $\times$ BS)}}}
& \multicolumn{3}{c}{\textbf{Video Quality}}
& \multicolumn{2}{c}{\textbf{Camera Error}}
& \multicolumn{3}{c}{\textbf{Semantic Consistency}} \\
\cmidrule(lr){3-5}
\cmidrule(lr){6-7}
\cmidrule(lr){8-10}

&
& FVD $\downarrow$
& FID $\downarrow$
& LPIPS $\downarrow$
& RotErr $\downarrow$
& TransErr $\downarrow$
& mIoU $\uparrow$
& $SR_e$ $\uparrow$
& $\mathrm{CLIP}_{\mathrm{local}}$ $\uparrow$ \\
\midrule

Direct last-frame SFT
& $4\mathrm{K}\times32=128\mathrm{K}$
& \third{114.03}
& \second{12.96}
& \second{0.43}
& \second{2.98}
& \second{0.56}
& \third{0.44}
& \second{0.56}
& \second{0.23} \\

D2S-SFT
& $4\mathrm{K}\times32=128\mathrm{K}$
& \second{110.53}
& \third{12.99}
& \first{0.42}
& \third{3.02}
& \first{0.51}
& \second{0.47}
& \first{0.59}
& \second{0.23} \\

\textbf{Ours}
& $500\times16=8\mathrm{K}$
& \first{99.35}
& \first{12.84}
& \first{0.42}
& \first{2.97}
& \third{0.59}
& \first{0.51}
& \first{0.59}
& \first{0.24} \\

\cmidrule(lr){1-10}

Ours w/o SFT anchor
& --
& 129.14
& 17.16
& 0.45
& 3.36
&  0.68
& 0.49
& 0.57
&  0.22 \\

\bottomrule
\end{tabular}
}
\vspace{-0.5em}
\label{tab:ablation_opsd}
\end{table*}

\begin{table*}[!t]
\centering
\small
\setlength{\tabcolsep}{4pt}
\caption{
\textbf{Ablation study of dual-mode OPSD.}
The dense-layout teacher and the student before OPSD training are included as references.
}
\vspace{-0.5em}
\resizebox{\textwidth}{!}{
\begin{tabular}{lccccccc|cccc}
\toprule
\multirow{2}{*}{\textbf{OPSD Variant}}
&
\multicolumn{7}{c|}{\textbf{Lastframe-Layout Mode}}
&
\multicolumn{4}{c}{\textbf{Camera-Only Mode}}
\\
\cmidrule(lr){2-8}
\cmidrule(lr){9-12}
&
FVD $\downarrow$
&
FID $\downarrow$
&
RotErr $\downarrow$
&
TransErr $\downarrow$
&
$\mathrm{mIoU}$ $\uparrow$
&
$SR_{e}$ $\uparrow$
&
$\mathrm{CLIP}_{\mathrm{local}}$ $\uparrow$
&
FVD $\downarrow$
&
FID $\downarrow$
&
RotErr $\downarrow$
&
TransErr $\downarrow$
\\
\midrule
\rowcolor{refgray}
Teacher (Dense layout)
& 123.50 & 13.72 & 2.92 & 0.51 & 0.60 & 0.60 & 0.23
& 147.20 & 14.25 & 3.85 & 0.63 \\

\rowcolor{refgray}
Student @ step0
& 137.10 & 14.08 & 3.58 & 0.58 & 0.40 & 0.56 & 0.22
& 147.20 & 14.25 & 3.85 & 0.63 \\

\cmidrule(lr){1-12}

Lastframe-layout single-mode
& \second{102.07} & \second{13.07} & \first{2.88} & \first{0.58} & \second{0.49} & \first{0.59} & \second{0.23}
& \third{121.88} & \second{13.42} & \second{3.09} & \third{0.67} \\

Camera-only single-mode
& \third{108.30} & \third{14.00} & \third{3.04} & \second{0.59} & \third{0.38} & \third{0.57} & \third{0.22}
& \second{105.22} & \third{13.77} & \first{3.03} & \first{0.63} \\

\textbf{Ours} (dual-mode)
& \first{99.35} & \first{12.84} & \second{2.97} & \second{0.59} & \first{0.51} & \first{0.59} & \first{0.24}
& \first{95.66} & \first{13.18} & \third{3.30} & \first{0.63} \\

\bottomrule
\end{tabular}
}
\vspace{-0.5em}
\label{tab:ablation_dual_mode}
\end{table*}

\begin{table*}[!t]
    \centering
    \small
    \setlength{\tabcolsep}{4pt}
    \caption{
    \textbf{Ablation study of mode sampling probability in dual-mode OPSD.}
    $p_{\mathrm{layout}}$ denotes the probability of sampling the lastframe-layout mode during training.
    }
    \vspace{-0.5em}
    \resizebox{\textwidth}{!}{
    \begin{tabular}{lccccccc|cccc}
    \toprule
    \multirow{2}{*}{$\mathbf{p_{\mathrm{layout}}}$}
    &
    \multicolumn{7}{c|}{\textbf{Lastframe-Layout Mode}}
    &
    \multicolumn{4}{c}{\textbf{Camera-Only Mode}}
    \\
    \cmidrule(lr){2-8}
    \cmidrule(lr){9-12}
    &
    FVD $\downarrow$
    &
    FID $\downarrow$
    &
    RotErr $\downarrow$
    &
    TransErr $\downarrow$
    &
    mIoU $\uparrow$
    &
    $SR_{e}$ $\uparrow$
    &
    $\mathrm{CLIP}_{\mathrm{local}}$ $\uparrow$
    &
    FVD $\downarrow$
    &
    FID $\downarrow$
    &
    RotErr $\downarrow$
    &
    TransErr $\downarrow$
    \\
    \midrule
    
    0.5
    & \third{102.63}
    & \third{13.40}
    & \first{2.936}
    & \second{0.620}
    & \third{0.4898}
    & \first{0.5862}
    & \third{0.2306}
    & \third{106.01}
    & \third{13.85}
    & \first{3.190}
    & \second{0.703} \\

    0.7
    & \second{99.35}
    & \first{12.84}
    & \second{2.968}
    & \first{0.593}
    & \first{0.5094}
    & \second{0.5861}
    & \first{0.2351}
    & \first{95.66}
    & \first{13.18}
    & \second{3.304}
    & \first{0.632} \\
    
    0.9
    & \first{93.05}
    & \second{12.96}
    & \third{2.978}
    & \third{0.653}
    & \second{0.5060}
    & \third{0.5832}
    & \second{0.2321}
    & \second{98.62}
    & \second{13.34}
    & \third{3.357}
    & \third{0.757} \\
    
    \bottomrule
    \end{tabular}
    }
    \vspace{-1.5em}
    \label{tab:ablation_mode_sampling}
\end{table*}

\textbf{SFT vs. OPSD.}
We compare OPSD with direct last-frame SFT and dense-to-sparse curriculum SFT (D2S-SFT) in \cref{tab:ablation_opsd} and \cref{fig:ablation_sft_vs_opsd}.
Directly optimizing the last-frame layout condition with SFT yields limited controllability.
Introducing a dense-to-sparse curriculum improves mIoU from 0.44 to 0.47, suggesting that this curriculum facilitates adaptation to sparse layout conditioning.
However, SFT still requires substantial optimization to adapt to the last-frame-only condition.
In contrast, our OPSD surpasses the SFT baselines in 5 out of 8 metrics using only $8$K training sample updates, compared with $128$K for the SFT baselines.
For simplicity, the training cost reported in \cref{tab:ablation_opsd} excludes the first 4K SFT steps for all three methods and measures only the subsequent adaptation cost. A full cost comparison, including the dense-layout SFT preceding OPSD, is provided in \cref{supp:implem}.
This demonstrates that OPSD provides a substantially more data-efficient and effective way to achieve the last-frame layout control.

\textbf{SFT Anchor Loss in OPSD.}
As shown in \cref{tab:ablation_opsd}, OPSD alone without the SFT anchor loss can effectively transfer privileged layout information, but may drift away from the original data distribution and degrade performance.

\textbf{Dual-Mode OPSD vs. Single-Mode OPSD.}
Our future-layout control is built upon camera control, and the two control modalities are therefore not fully independent.
As shown in \cref{tab:ablation_dual_mode}, lastframe-layout single-mode OPSD also improves camera-only inference over the step-0 student.
Camera-only single-mode OPSD improves camera-only performance compared to lastframe-layout single-mode OPSD, but degrades layout controllability  under the lastframe-layout inference setting.
In contrast, our dual-mode OPSD jointly distills both modes from the same dense-layout teacher and achieves the best overall performance across the two inference settings.
It delivers the best video quality and layout controllability while maintaining comparable camera accuracy, suggesting that dual-mode training promotes beneficial interaction between camera and layout conditioning.

\textbf{Mode Sampling Probability.}
We further study the sampling probability between the lastframe-layout and camera-only modes in \cref{tab:ablation_mode_sampling}.
$p_{\mathrm{layout}}=0.7$ achieves the best overall results, so we use it as our default setting and in other experiments.

More ablation studies are included in \cref{supp-sec:abl}.

\section{Related Work}

\subsection{Controllable Video Generation}
\textbf{Camera Control.}
Camera-controllable video generation~\citep{bahmani2025vd3d,bai2025recammaster,bai2024syncammaster,zheng2024cami2v,li2025realcam,yu2025trajectorycrafter} aims to explicitly control viewpoint trajectory during synthesis.
Recent works use camera extrinsics~\citep{wang2024motionctrl,bai2025recammaster}, Pl\"ucker-ray embeddings~\citep{he2024cameractrl,bahmani2025ac3d,he2025cameractrl2}, or explicit 3D priors~\citep{ren2025gen3c,cao2025uni3c,wang2025epic} as camera representations and inject them into video diffusion models.
Some approaches also incorporate relative camera geometry into attention through positional encodings~\citep{Zhang_2026_UCPE,li2026rerope}.
More recently, world models extend camera control towards long-horizon scene exploration~\citep{li2025hunyuangamecraft,sun2026hunyuan1_5,mao2025yume15,robbyantteam2026lingbotworld}.
However, these methods only determine how the viewpoint should move.
Our work complements camera control with object-level layout guidance, enabling explicit control of the semantic and spatial composition of future views beyond the initially observed regions.

\textbf{Object Motion and Layout Control.}
Spatiotemporal control has been extensively studied for manipulating object motion. Video generation models use point trajectories~\citep{geng2024motionprompting,wang2024levitor,diffusionasshader}, bounding boxes~\citep{jain2024peekaboo,wu2024motionbooth,wang2025cinemaster,li2025magicmotion}, or masks~\citep{wu2024draganything,Yariv_2025_throughmask} to specify object motion over time.
Some works can also jointly control object motion and camera~\citep{yang2024directavideo,wu2024motionbooth,chen2025perception,xing2025motioncanvas,zheng2026versecrafter}.
These methods specify how an object moves, but typically assume that the controlled object is already present in the first frame.
Relatedly, layout-conditioned generation specifies what objects should appear and where. While it has been widely explored in image generation~\citep{wang2024instancediffusion,zhang2025eligen,zhang2025creatilayout,huang2025laytrol}, it remains less explored in video generation. 
Existing methods rely on dense per-frame object descriptions~\citep{li2025trackdiffusion,feng2025blobgenvid}, mainly targeting object motion.
In contrast, \ourmodel targets future-view layout control under large viewpoint changes, supporting last-frame-only layout conditioning.

\subsection{On-Policy Self-Distillation}
On-policy distillation (OPD)~\citep{Agarwal2024opdllm} has recently emerged as an effective alternative post-training method complementary to SFT and RL.
OPD trains a student to match a teacher on trajectories generated by the student itself, thereby reducing the train–inference distribution mismatch of SFT.
Compared with reinforcement learning with verifiable rewards (RLVR) such as GRPO~\citep{shao2024grpo}, which typically relies on a scalar sequence-level reward, OPD offers dense token-level supervision from the teacher.
Recent works~\citep{fang2026flowopd, li2026diffusionopd,zhou2026danceopd, xu2026qwenimage20rl,fu2026anyopd} extend this method to diffusion and flow-matching models. 
On-policy self-distillation (OPSD)~\citep{zhao2026opsdllm,jiang2026dopsd} further uses the model itself as the teacher by providing it with richer context, known as privileged information, while the student receives only inference-time conditions.
It removes the need for a separately trained or larger teacher.
While OPSD has recently received increasing attention in LLMs, it remains relatively underexplored in image and video generation~\citep{li2026rethinkingclassifierfreeguidanceonpolicy, liu2026opsdv}.
\ourmodel uses OPSD to achieve last-frame layout control and encourage the synergy between camera and layout conditioning.

\section{Conclusion}

In this paper, we introduce \ourmodel, a unified framework for future-view layout control under large viewpoint changes.
\ourmodel enables users to control both camera motion and the semantic and spatial composition of newly revealed regions through two inference modes: a single mode conditioned on the camera trajectory and a dual mode additionally conditioned on the last-frame layout.
To support this setting, we curate LIFT-Vista, a dataset featuring substantial future-region revelation with temporally consistent camera and layout annotations.
To effectively learn from sparse last-frame layout guidance, we adopt on-policy self-distillation (OPSD) and apply it in a dual-mode training scheme, transferring privileged dense-layout knowledge to a shared student across both inference modes.
Together, these designs improve camera and future-layout controllability within a unified video generation framework. 

\noindent \textbf{Acknowledgement.} This work is supported by U.S. National Science Foundation Award IIS-2433768 and IIS-2127544. This work also used DeltaAI at National Center for Supercomputing Applications (NCSA) through allocation CIS260420 from the Advanced Cyberinfrastructure Coordination Ecosystem: Services \& Support (ACCESS) program, which is supported by U.S. National Science Foundation grants \#2138259, \#2138286, \#2138307, \#2137603, and \#2138296.

\bibliography{iclr2027_conference}
\bibliographystyle{iclr2027_conference}

\newpage

\appendix
\section{Appendix}
\label{sec:supp}

\subsection{Data Curation Details}
\label{supp-sec:data-curation}
We provide additional details of the data curation pipeline described in \cref{sec:data-curation} and \cref{fig:data-curation}.

\textbf{Metadata Filtering and Clip Extraction.}
We curate our dataset from SpatialVID-HQ~\citep{wang2025spatialvid}, Sekai-HQ~\citep{li2026sekai}, and RealEstate10K~\citep{zhou2018realestate10k}.
For SpatialVID and Sekai, we first remove crowded scenes using the
scene metadata, since dense crowds often lead to ambiguous object
correspondence and noisy layout annotations.
For SpatialVID, we additionally remove broad natural-landscape categories that contain few meaningful foreground objects for layout control.
For RealEstate10K, we retain only clips with at least 16 FPS, a duration of at least 5 seconds, and a resolution no smaller than $352 \times 640$.
Then, we extract candidate windows from source videos.
To match our training configuration, each source video is resampled to 16\,fps and cut into 81-frame windows ($\approx 5$\,s) with a stride of 40 frames.

\textbf{Camera-Motion Filtering.}
We compute the FoV expansion ratio and accumulated translation distance
using $K=8$ uniformly sampled keyframes.
For $r_{\mathrm{FoV}}$, we use $50{,}000$ directions sampled on the unit sphere; a direction is visible in a keyframe if it projects inside the image with positive depth under that frame's pinhole camera.
For intuition, \(r_{\mathrm{FoV}}=1\) indicates no expansion of angular viewing coverage, while \(r_{\mathrm{FoV}}\approx1.5\), \(2.0\), and \(3.0\) roughly correspond to \(45^\circ\), \(90^\circ\), and \(180^\circ\) yaw rotations, respectively, for a $90^\circ$ horizontal FoV.
Because the original camera annotations from different source datasets
follow different translation scales, we use dataset-specific translation
thresholds during the initial filtering stage.
Specifically, candidate windows are retained if they satisfy either the
FoV or translation criterion:
\[
\begin{array}{c|cc}
\text{Dataset} & r_{\mathrm{FoV}} & d_{\mathrm{trans}} \\
\hline
\text{SpatialVID} & \geq 1.4 & \geq 2.0 \\
\text{Sekai} & \geq 1.4 & \geq 0.16 \\
\text{RealEstate10K} & \geq 1.3 & \geq 10.0
\end{array}
\]
To reduce highly redundant windows from long source videos and keep diversity, we further rank candidates with $r_{FoV}$ in decreasing order, apply temporal non-maximum suppression, and retain at most two non-overlapping windows with the highest $r_{\mathrm{FoV}}$ from each source video.

\textbf{Content-Change Filtering.}
To better match our target setting of future-view layout control, we further apply content-change filtering to focus on clips with substantial future-region revelation for Stage 2 and 3 training, where the camera motion exposes content that is not visible in the first frame. 
The content change ratios are computed between the first and last frames with DINOv2-giant~\citep{oquab2023dinov2}. We resize the frames without cropping while preserving their aspect ratio so that the longer side is approximately 574\, px (a multiple of the 14\, px patch size).
For every patch in one frame, we search for its most similar patch in the other frame and use a cosine-similarity threshold of $\tau=0.5$.
We compute both directions, corresponding to newly revealed content in the last frame ($r_{lf\text{-}CCR}$) and content leaving the first frame ($r_{ff\text{-}CCR}$).
We remove clips with $r_{\mathrm{FoV}} < 1.5$, $r_{lf-CCR} < 0.1$ and $r_{ff-CCR} < 0.1$, i.e.\ clips with limited expansion of angular viewing coverage whose content is nearly unchanged in both directions.

\textbf{Layout Annotation.}
We annotate object layouts in the last frame using Qwen3-VL-32B~\citep{bai2025qwen3vl}, which we find produces more reliable and selective annotations of meaningful foreground objects.
We prompt the model to select salient and spatially meaningful foreground instances while excluding tiny clutter, background regions, severely occluded objects, and excessively large regions.
Clips with no valid candidates are discarded.

\begin{figure*}[t]
    \centering
    \vspace{-0.1cm}
    \includegraphics[width=\linewidth]{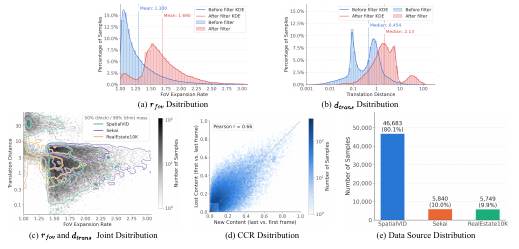}
    \vspace{-0.4cm}
    \caption{
    \textbf{Visualization of data distribution.} (a, b) Distributions of the camera-motion metrics over
  all candidate windows before filtering (blue) and over the final layout training set (red). Filtering removes the mass of near-static windows and shifts both metrics toward larger viewpoint changes.
  (c) Joint distribution of the two camera-motion metrics in the training set, with 50\% and 90\% mass contours per source dataset.
  (d) Joint distribution of the two content change ratios. The two directions are correlated and complementary but not redundant.
  (e) Number of training clips per dataset.
    }
    \label{fig:data_distribution}
\end{figure*}

\textbf{Data Statistics.} 
After curation, we obtain 120,898 training samples for Stage 1, a subset of 58,272 samples for Stages 2 and 3, and 600 test samples.
The test set is randomly sampled from the curated data according to three FoV-expansion ranges, $[1.0,1.5)$, $[1.5,2.0)$, and $[2.0,\infty)$, with a sampling ratio of $1{:}2{:}2$, so as to cover different levels of viewpoint change.
All test samples are excluded from the training sets.
The distribution of the curated dataset is visualized in \cref{fig:data_distribution}.
The resulting layout annotations contain an average of 5.5 objects per clip and cover $8,302$ categories, including indoor objects (chair, window, lamp, cabinet, sofa, etc.) and outdoor objects (person, building, car, boat, tree, sign, etc.).
$84.8\%$ of clips contain objects that are invisible in the first frame but appear in future views, directly supporting our future-view layout-control setting.

\subsection{More Experiments}
\subsubsection{More Implementation Details}
\label{supp:implem}
We summarize the training configuration of the three stages in
\cref{tab:training_config}.
Stage 1 establishes camera controllability, Stage 2 learns dense
spatiotemporal layout control, and Stage 3 transfers the dense-layout
capability to last-frame-layout and camera-only modes through dual-mode
OPSD.
All stages are trained using AdamW on 4 NVIDIA H100 GPUs at a resolution
of $352\times640$ with 81-frame clips at 16 FPS.

\begin{table}[t]
    \centering
    \small
    \setlength{\tabcolsep}{7pt}
    \caption{\textbf{Training configuration of \ourmodel across the three stages.}}
    \label{tab:training_config}
    \begin{tabular}{lccc}
    \toprule
    \textbf{Setting}
    & \textbf{Stage 1}
    & \textbf{Stage 2}
    & \textbf{Stage 3} \\
    \midrule
    
    Training mode
    & Camera Control (SFT)
    & Dense Layout Control (SFT)
    & Dual-Mode OPSD \\

    Base Model
    & \makecell[c]{Wan2.1-Fun-V1.1-1.3B-\\Control-Camera}
    & Stage 1
    & Stage 2 \\

    Dataset size
    & 120,898
    & 58,272
    & 58,272 \\
    
    Training steps
    & 8,000
    & 4,000
    & 500 \\
    
    Global batch size
    & 32
    & 32
    & 16 \\
    
    Learning rate
    & $1\times10^{-5}$
    & $1\times10^{-4}$
    & $5\times10^{-5}$ \\
    
    Optimizer
    & AdamW
    & AdamW
    & AdamW \\
    
    \bottomrule
    \end{tabular}
\end{table}

\subsubsection{Evaluation Metrics.}
We evaluate the generated videos along three dimensions: visual quality, camera controllability, and layout controllability.

\textbf{Visual quality.}
We report Fr\'echet Video Distance (FVD)~\citep{unterthiner2018towards},
Fr\'echet Inception Distance (FID)~\citep{heusel2017gans}, and
Learned Perceptual Image Patch Similarity (LPIPS)~\citep{zhang2018unreasonable}.
FVD measures the distributional discrepancy between generated and real videos in a learned video feature space, jointly reflecting visual realism and temporal coherence.
FID measures the distributional similarity between generated and real frames in the image feature space and primarily evaluates frame-level visual fidelity.
LPIPS measures the perceptual distance between generated and corresponding reference frames using deep visual features.
Lower values indicate better performance for all three metrics.

\textbf{Camera controllability.}
Following ~\citep{zhang2025dualcamctrl},
we evaluate camera trajectory accuracy using rotation error (RotErr) and translation error (TransErr).
RotErr measures the angular discrepancy between the estimated camera rotations of the generated video and the target camera trajectory, while TransErr measures their translation discrepancy. We use the off-the-shelf model~\citep{lin2025depth} to estimate the camera pose of generated videos and then compare with the ground truth.
Lower RotErr and TransErr indicate more accurate adherence to the prescribed camera motion.

\textbf{Layout controllability.}
Following OverLayBench~\citep{li2026overlaybench}, we evaluate object-level spatial and semantic control using mean Intersection over Union (mIoU), Success Rate of Entity (SR$_e$), and CLIP$_{\mathrm{local}}$~\citep{radford2021learning}. We use Qwen3.6-27B~\citep{bai2025qwen3vl} to detect objects and their locations in the generated videos.
mIoU measures the spatial overlap between generated object regions and their target layout boxes, reflecting object placement accuracy.
SR$_e$ measures the fraction of conditioned entities that are successfully generated with the intended semantics and spatial placement.
CLIP$_{\mathrm{local}}$ computes the CLIP similarity between local regions corresponding to conditioned objects and their text descriptions, measuring local semantic consistency.
Higher mIoU, SR$_e$, and CLIP$_{\mathrm{local}}$ indicate better layout controllability.

\begin{table*}[!t]
\centering
\small
\setlength{\tabcolsep}{4.5pt}
\caption{
\textbf{Comparisons of different layout sparsity.}
We compare four layout sparsities without retraining: layou bboxes on all 81 frames, on 8 frames (10, 20, \dots, 70, 81), on 4 frames (20, 40, 60, 81), and on the last frame only.
$^\dagger$ marks our target sparsity.
}
\vspace{-0.2cm}
\resizebox{\textwidth}{!}{
\begin{tabular}{lccccccccc}
\toprule
\multirow{2}{*}{\textbf{Layout given at}}
& \multicolumn{3}{c}{\textbf{Video Quality}}
& \multicolumn{2}{c}{\textbf{Camera Error}}
& \multicolumn{3}{c}{\textbf{Semantic Consistency}}
\\
\cmidrule(lr){2-4}
\cmidrule(lr){5-6}
\cmidrule(lr){7-9}
& FVD $\downarrow$
& FID $\downarrow$
& LPIPS $\downarrow$
& RotErr $\downarrow$
& TransErr $\downarrow$
& mIoU $\uparrow$
& $SR_e$ $\uparrow$
& $\mathrm{CLIP}_{\mathrm{local}}$ $\uparrow$
\\
\midrule

Dense
& 98.84 & \textbf{12.58} & \textbf{0.398} & \textbf{2.544} & 0.594 & \textbf{0.6217} & \textbf{0.6081} & \textbf{0.2403} \\

8 frames
& \textbf{92.72} & 12.86 & \underline{0.406} & 2.772 & \underline{0.574} & 0.5758 & \underline{0.5907} & \underline{0.2384} \\

4 frames
& \underline{95.86} & \underline{12.79} & 0.407 & \underline{2.751} & \textbf{0.559} & \underline{0.5764} & 0.5859 & 0.2379 \\

Last frame only\,$^\dagger$
& 99.35 & 12.84 & 0.418 & 2.968 & 0.593 & 0.5094 & 0.5861 & 0.2351 \\

\bottomrule
\end{tabular}
}
\label{tab:any_sparse}
\end{table*}

\begin{table}[t]
    \centering
    \small
    \setlength{\tabcolsep}{5pt}
    \caption{
    \textbf{Implementation details for the SFT and OPSD ablations.}
    Direct LF-SFT denotes direct last-frame SFT.
    D2S-SFT denotes dense-to-sparse curriculum SFT.
    For a fair comparison, all three methods are counted from the same Stage~1 camera-control checkpoint, and the dense-layout SFT stage required before OPSD is included in the training cost of our method.
    Sample updates are computed as training steps $\times$ global batch size over the full adaptation process.
    }    
    \resizebox{\linewidth}{!}{
    \begin{tabular}{lccc}
    \toprule
    \textbf{Setting}
    & \textbf{Direct LF-SFT}
    & \textbf{D2S-SFT}
    & \textbf{OPSD (Ours)} \\
    \midrule

    Starting checkpoint
    & Stage 1
    & Stage 1
    & Stage 1 \\

    Training schedule
    & 8K LF
    & 4K Dense + 2K 8-frame + 2K LF
    & 4K Dense + 500 OPSD \\

    Global batch size
    & 32
    & 32
    & 32 (Dense) / 16 (OPSD) \\

    Learning rate
    & $1\times10^{-4}$
    & $1\times10^{-4}$
    & $1\times10^{-4}$ (Dense) / $5\times10^{-5}$ (OPSD) \\

    Total training steps
    & 8,000
    & 8,000
    & 4,500 \\

    Sample updates (Steps $\times$ BS)
    & 256K
    & 256K
    & 136K \\

    Dataset size in pool
    & 58,272
    & 58,272
    & 58,272 \\

    Optimizer
    & AdamW
    & AdamW
    & AdamW \\

    \bottomrule
    \end{tabular}
    }
    \label{tab:training_cost_sft_vs_opsd}
\end{table}

\begin{figure*}[t]
    \centering
    \includegraphics[width=\linewidth]{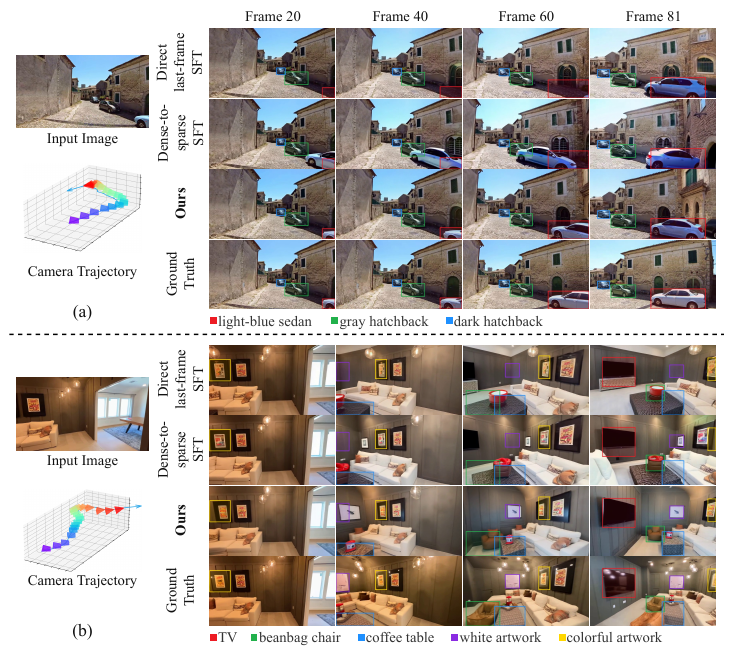}
    \caption{\textbf{SFT and OPSD Comparison.}
    (a) Direct last-frame SFT largely ignores the sparse layout condition, leaving the target sedan absent until the final frame, while dense-to-sparse SFT introduces it too early and with inaccurate spatial alignment. In contrast, OPSD produces a trajectory that better matches the ground-truth evolution.
    (b) Direct last-frame SFT fails to realize several conditioned objects, whereas dense-to-sparse SFT improves object presence but still exhibits poor temporal alignment before the last frame. OPSD more faithfully follows both the target layout and its temporal evolution.
    }
    \label{fig:ablation_sft_vs_opsd}
\end{figure*}

\begin{figure*}[t]
    \centering
    \includegraphics[width=\linewidth]{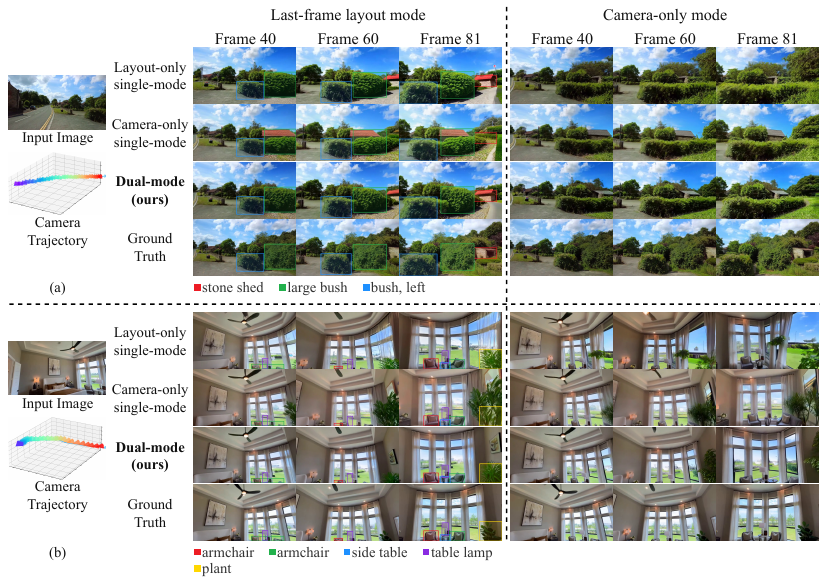}
    \caption{\textbf{Dual-mode OPSD Comparison.} 
    The last-frame-layout mode uses both the camera trajectory and last-frame layout, whereas the camera-only mode uses only the camera trajectory.
    Last-frame-layout single-mode OPSD follows the prescribed layout, but yields lower video quality than dual-mode OPSD in camera-only inference, while camera-only single-mode OPSD preserves camera control but often fails to realize the specified future layout.
    In contrast, dual-mode OPSD maintains strong camera control in both inference settings while faithfully following the future-layout condition when provided.
    }
    \label{fig:ablation_dual_mode}
\end{figure*}

\begin{figure*}[!t]
    \centering
    \includegraphics[width=\linewidth]{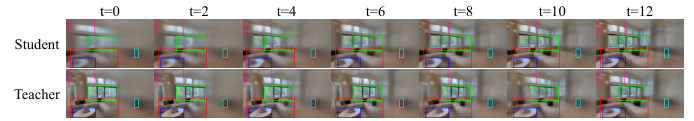}
    \caption{\textbf{Student--teacher comparison} in dense-to-lastframe layout OPSD.
    The labels $t=0,2,\ldots,12$ denote denoising-step indices rather than the continuous diffusion time used in the equations.
    Within the first 10 denoising steps, the dense-layout teacher provides substantial corrections on the student-visited states, producing results that better conform to the specified layout.
    Afterwards, the teacher correction becomes much weaker, motivating us to concentrate OPSD supervision on the first 10 high-noise states.
    }
    \label{fig:state_select}
\end{figure*}

\subsubsection{More Ablation Studies}
\label{supp-sec:abl}

\textbf{Any Layout Sparsity.}
As shown in \cref{tab:any_sparse}, \ourmodel supports different layout sparsity at inference without retraining.
Providing more layout frames generally improves generation quality, spatial alignment and semantic consistency.
These results demonstrate that \ourmodel can leverage additional layout guidance when available, allowing users to trade annotation effort for finer spatial control while retaining the practical last-frame-only interface.

\textbf{SFT vs. OPSD.}
A more detailed training cost comparison between SFT and OPSD is illustrated in \cref{tab:training_cost_sft_vs_opsd}. OPSD is substantially more data-efficient and effective at achieving the last-frame layout control. 
The qualitative comparisons in \cref{fig:ablation_sft_vs_opsd} illustrate the layout-following limitations of the evaluated SFT baselines in our setting.
Because the last-frame layout provides only a weak endpoint signal, direct last-frame SFT often ignores conditioned objects.
Dense-to-sparse SFT partially alleviates this issue, but the model still has to infer, from the last-frame layout alone, how the specified objects should emerge and evolve throughout the preceding frames under camera motion.
In contrast, OPSD provides direct supervision from a dense-layout teacher on the student's own rollout states, supplying explicit guidance for the intermediate scene evolution and making the sparse future-layout condition substantially easier to learn.

\textbf{Dual-mode OPSD.}
The qualitative comparisons are shown in \cref{fig:ablation_dual_mode}.
Lastframe-layout single-mode OPSD learns to follow the future layout, but specializing only to this conditioning mode can yield lower visual quality than dual-mode OPSD under camera-only inference, e.g., causing scene drift or geometric distortion.
Conversely, camera-only single-mode OPSD maintains strong camera-conditioned generation but lacks sufficient supervision for future-layout control, often producing incorrect objects or failing to place them at the specified locations.
By alternating between both modes and distilling from the same dense-layout teacher, dual-mode OPSD preserves camera controllability while retaining accurate future-layout control, leading to more consistent behavior across both inference settings.

\begin{figure*}[!t]
    \centering
    \vspace{-0.1cm}
    \includegraphics[width=\linewidth]{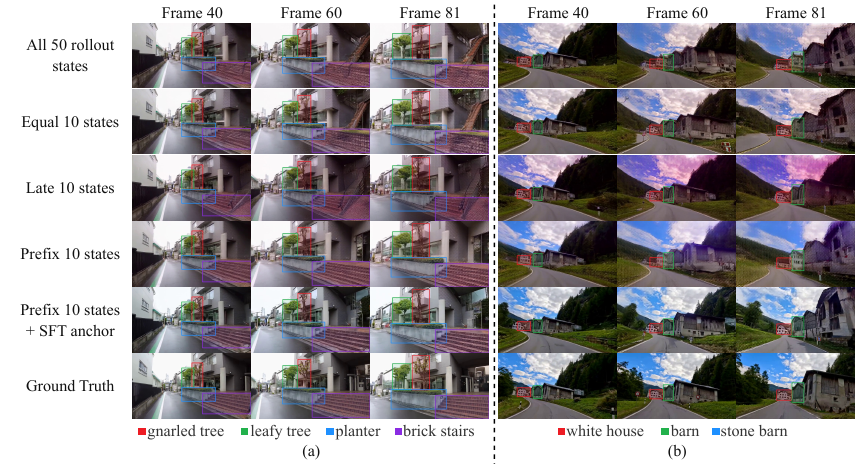}
    \vspace{-2em}
    \caption{\textbf{Comparison of different ODE state sampling in OPSD.} 
    Distilling only on the late 10 states (the third row) yields the weakest layout following, indicating that global layout structure is mainly determined during the early high-noise stage.
    Prefix-state distillation provides stronger layout control, while adding the SFT anchor further prevents drifting and visual-quality degradation.
    }
    \label{fig:ablation_state}
\end{figure*}

\textbf{ODE State Sampling Strategy in OPSD.}
We study where along the ODE trajectory the teacher provides the most effective distillation signal.
As visualized in \cref{fig:state_select}, during the early high-noise stage, the dense-layout teacher produces clear corrections to the student-visited state, especially in the global object layout and spatial arrangement.
In contrast, after roughly the first 10 denoising steps, the student and teacher predictions become much closer, and the teacher correction is significantly weaker and provides less informative distillation signals.
As shown in \cref{tab:ablation_state_selection}, prefix-state distillation provides a favorable trade-off between training time, camera accuracy, and spatial layout control.
Based on this observation, we select the first 10 high-noise rollout states for OPSD.
This selective state sampling focuses optimization on the phase where the global layout structure is established while reducing the cost of student rollout and distillation.
This ablation is conducted on lastframe-layout single-mode OPSD. 
Distilling only the last 10 states performs substantially worse across all quality and controllability metrics, indicating that low-noise states provide little useful signal for correcting the global layout.
In contrast, supervising the prefix 10 high-noise states yields substantially better camera and layout control.
Combined with the SFT anchor, prefix-10-state distillation achieves the strongest overall performance while introducing little computation overhead.
This is substantially more efficient than querying all 50 rollout states.
These results support our observation that the dense-layout teacher provides its most informative corrections during the early high-noise stage, where the global scene and layout structure are primarily established. Visualization comparisons are shown in \cref{fig:ablation_state}.

\begin{figure}[!t]
    \centering
    \includegraphics[width=0.8\linewidth]{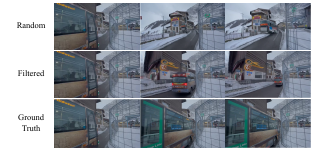}
    \vspace{-0.2cm}
    \caption{
\textbf{Comparison of the proposed camera-motion filtering strategy.}
The first row shows results from the model trained on randomly sampled clips,
exhibiting degraded visual quality and less stable object appearance.
The second row shows results from the model trained on filtered dynamic clips,
with better preservation of scene structure and object appearance under camera motion.
The third row shows ground-truth frames.
}
    \label{fig:data_filter_ablation}
\end{figure}

\textbf{Data Filtering.} 
Table.~\ref{tab:ablation_data_filter} and Fig.~\ref{fig:data_filter_ablation} demonstrate the effectiveness of our camera-motion filtering strategy.
For a fair comparison, we construct two equally sized training sets of 30K clips, one using our proposed filtering strategy and the other using random sampling.
We additionally hold out a test set of 250 clips featuring substantial future-region revelation.
Compared with random sampling, our filtered training data consistently improves all evaluation metrics, including video quality and camera-control accuracy, and better preserves scene structure and object appearance under large viewpoint changes.
These results highlight the importance of curating training data with large viewpoint changes for our target setting.

\begin{table*}[!t]
\centering
\small
\setlength{\tabcolsep}{4.5pt}
\caption{\textbf{Ablation of OPSD state selection.}}
\resizebox{\textwidth}{!}{
\begin{tabular}{lccccccccc}
\toprule
\multirow{2}{*}{\textbf{Method}}
& \multicolumn{3}{c}{\textbf{Video Quality}}
& \multicolumn{2}{c}{\textbf{Camera Error}}
& \multicolumn{3}{c}{\textbf{Semantic Consistency}} 
& \textbf{Time} \\

\cmidrule(lr){2-4}
\cmidrule(lr){5-6}
\cmidrule(lr){7-9}
\cmidrule(lr){10-10}
& FVD $\downarrow$
& FID $\downarrow$
& LPIPS $\downarrow$
& RotErr $\downarrow$
& TransErr $\downarrow$
& mIoU $\uparrow$
& $SR_e$ $\uparrow$
& $\mathrm{CLIP}_{\mathrm{local}}$ $\uparrow$ 
& s/step $\downarrow$ \\
\midrule

All 50 rollout states
& 127.24
& \underline{15.02}
& 0.441
& 3.467
& 0.624
& 0.4786
& 0.5740
& \underline{0.2234}
& 544.82 \\

Equal 10 states
& \underline{120.23}
& 15.17
& \underline{0.439}
& 3.228
& 0.589
& 0.4807
& \underline{0.5777}
& 0.2223
& 185.28 \\

Late 10 states
& 197.63
& 17.82
& 0.494
& 6.347
& 0.906
& 0.3089
& 0.5227
& 0.2024
& 183.56 \\

Prefix 10 states
& 127.02
& 17.77
& 0.442
& \underline{2.961}
& \textbf{0.544}
& \underline{0.4922}
& 0.5664
& 0.2219
& \textbf{117.21} \\

Prefix 10 states + SFT anchor
& \textbf{102.07}
& \textbf{13.07}
& \textbf{0.422}
& \textbf{2.879}
& \underline{0.583}
& \textbf{0.4934}
& \textbf{0.5883}
& \textbf{0.2313}
& \underline{125.20} \\

\bottomrule
\end{tabular}
}

\label{tab:ablation_state_selection}
\end{table*}

\begin{table}[!t]
\centering
\small
\caption{\textbf{Ablation of the proposed camera-motion filtering strategy.}
We compare models trained on randomly sampled and filtered data under the same training configuration.
The proposed filtering strategy consistently improves both visual quality and camera-control accuracy.}
\begin{tabular}{lccccc}
\toprule
Method & FVD $\downarrow$ & FID $\downarrow$ & LPIPS $\downarrow$ & RotErr $\downarrow$ & TransErr $\downarrow$ \\
\midrule
w/o filter   & 278.85 & 32.88 & 0.52 & 15.62 & 4.05 \\
w/ filter & \textbf{266.58} & \textbf{30.99} & \textbf{0.51} & \textbf{13.14} & \textbf{3.87} \\
\bottomrule
\end{tabular}
\label{tab:ablation_data_filter}
\end{table}

\subsubsection{User Study}
\label{supp-sec:user_study}
To complement automatic evaluation, we conduct a randomized user study with 7 participants.
For each trial, participants are shown the input condition and anonymized generated videos from different methods in random order, and are asked to choose the result that best satisfies the control intent, considering camera motion, layout adherence, object consistency, and visual quality.
We compare with the best two methods in \cref{tab:quantitative_comparison} in this user study.
Each participant evaluates 30 comparisons, yielding 210 responses in total.
Participants may also select a ``None preferred'' option, and the reported preference rates are computed over responses that select one of the three methods.
As shown in \cref{tab:user_study}, our method achieves the highest preference rate, 65.96\%.
This result indicates that the advantages of our method are not only reflected in automatic metrics but are also perceptible to human observers.
The user study further confirms that our approach produces visually convincing and controllable videos that better align with user intent.

\begin{table}[!t]
\centering
\small
\caption{
\textbf{User study }on preference rates among different camera-controllable video generation methods.
Participants select the result that best satisfies interactive editing requirements.
}
\begin{tabular}{lccc}
\toprule
 & Ours & Uni3C & MagicMotion \\
\midrule
Preference & \textbf{65.96\%} & 22.34\% & 11.70\% \\
\bottomrule
\end{tabular}
\label{tab:user_study}
\end{table}

\subsection{Limitation}
\ourmodel currently uses 2D bounding boxes with local text prompts to specify the desired future-view composition. Although this representation is simple and user-friendly, it provides only coarse spatial constraints and does not explicitly capture depth, orientation, or occlusion relationships between objects. Extending the layout representation with richer geometric or instance-level controls could enable more fine-grained specification of future views.

\end{document}